\documentclass{article}
\usepackage{graphicx} 
\usepackage{subcaption}
\usepackage{booktabs}
\usepackage{rotating}
\usepackage[utf8]{inputenc}
\usepackage{amsmath}
\usepackage{svg}
\usepackage{hyperref}
\hypersetup{
    colorlinks,
    citecolor=black,
    filecolor=black,
    linkcolor=black,
    urlcolor=black
}
\usepackage{tabularx}
\usepackage{adjustbox}
\usepackage{algorithm}
\usepackage{algorithmicx}  
\usepackage{algpseudocode} 
\usepackage{xcolor}
\usepackage{parskip}
\usepackage[final]{neurips_2024}

\begin{document}

\title{GEM-T: Generative Tabular Data via Fitting Moments}
\author{Miao Li$^{\ddagger}$, Phuc Nguyen$^{\ddagger}$, Christopher Tam$^{\ddagger}$, Alexandra Morgan, Kenneth Ge, \\ \textbf{Rahul Bansal, Linzi Yu, Rima Arnaout$^{\S}$, Ramy Arnaout$^{\S}$}
\thanks{Kenneth Ge, Miao Li, Alexandra Morgan, Phuc Nguyen, Rahul Bansal, Christopher Tam and Linzi Yu are/were with the Department of Pathology at Beth Israel Deaconess Medical Center (BIDMC), Boston, MA 02215 during the completion of this work. Rima Arnaout is with the Department of Medicine, the Bakar Computational Health Sciences Institute, and the UCSF UC Berkeley Joint Program for Computational Precision Health at the University of California San Francisco, San Francisco, CA 94143. Ramy Arnaout (to whom correspondence should be addressed at rarnaout@bidmc.harvard.edu) is with the Department of Pathology and the Division of Clinical Informatics, Department of Medicine, BIDMC and with Harvard Medical School, Boston, MA 02215.}
}

\maketitle
\begin{center}
$^{\ddagger}$ These authors contributed equally.\\
$^{\S}$ Joint senior authors.
\end{center}

\tableofcontents

\begin{abstract}
Tabular data dominates data science but poses challenges for generative models, especially when the data is limited or sensitive. We present a novel approach to generating synthetic tabular data based on the principle of maximum entropy---MaxEnt---called GEM-T, for ``generative entropy maximization for tables.'' GEM-T directly captures $n^{\text{th}}$-order interactions---pairwise, third-order, etc.---among columns of training data. In extensive testing, GEM-T matches or exceeds deep neural network approaches previously regarded as state-of-the-art in 23 of 34 publicly available datasets representing diverse subject domains (68\%). Notably, GEM-T involves orders-of-magnitude fewer trainable parameters, demonstrating that much of the information in real-world data resides in low-dimensional, potentially human-interpretable correlations, provided that the input data is appropriately transformed first. Furthermore, MaxEnt better handles heterogeneous data types (continuous vs. discrete vs. categorical), lack of local structure, and other features of tabular data. GEM-T represents a promising direction for light-weight high-performance generative models for structured data.
\end{abstract}

\section{Introduction}

Generative modeling has emerged as an important option for addressing privacy, data availability, and cost constraints across a range of data modalities and use cases~\cite{giuffre2023harnessing, assefa2020generating}. Recent breakthroughs have focused on text (large language models~\cite{vaswani2017attention, kaplan2020scaling}) and images (diffusion models \cite{ho2020denoising}), both alone and in combination (e.g., multimodal models such as CLIP~\cite{radford2021learning}). However, many high-impact datasets across many fields---including medicine, biology, communications, finance, agriculture, and industry---are neither text- nor image-based but tabular, where datapoints are rows and are described by features arranged as columns ~\cite{arnaout_pregnancy_2019,reddy_improving_2024,behr_learning_2024,subramaniam_ontology-guided_2025,panahiazar_gender-based_2022,arnaout_visualizing_2022}. Modelling tabular data remains particularly challenging because of the heterogeneity of data types, which can be numerical, categorical, or binary, and by the lack of consensus over how best to evaluate synthetic tabular data, especially considering the variety of research goals and downstream tasks \cite{grade_inflation, chundawat2022universal}.

Current approaches for generating synthetic tabular data~\cite{fonseca2023tabular,bauer2024comprehensive} (Section~\ref{sec:related}) leave much room for improvement. For example, consider copulas \cite{charpentierEstimationCopulasTheory2006}, which are mathematically appealing thanks to a result (Sklar's theorem) that allows for a separation between inter-column dependencies and the distribution of values within each column (the marginals), but do not work well in high dimensions (prompting remedies such as vines \cite{bedford2002vines}). The Gaussian copula in particular does not capture tail behavior well (it has zero tail-dependence); it is not always obvious what alternatives to choose (another elliptical copula such as the Student's t-copula, an Archimedean copula, or an extreme-value copula) to model both tails and the main distributions of the data. Meanwhile, despite the success of generative adversarial networks (GANs) in several settings, they often seem to struggle on tabular data \cite{pathare2023comparison}.

Deep neural networks (DNNs) have been highly successful in many settings in part because the large numbers of parameters they include enable them to fit very complicated distributions of data. However, this comes at the expense of generally requiring large datasets for training, as well as the risk of overfitting or indeed memorizing the input data. These can be important limitations in the context of generating synthetic tabular data, for which a common goal is privacy: to generate new data that captures and recapitulates dependencies in the training data while explicitly avoiding reproduction of the (private) training data. Concurrently, it is often the case that data lies on relatively low-dimensional manifolds that in principle should require many fewer parameters than DNNs typically use; i.e., datasets may often be simpler than they appear \cite{schmidt_distilling_2009, raissi2017physics}. Indeed, the above limitations of large models, as well as others (training time, storage, energy use), have created great interest in developing high-performance lightweight models (via binarization, quantization, pruning, retraining of small models on the output of larger reference models, etc.) as well as models that can train well on small datasets.

In this work, we propose maximum entropy (MaxEnt)~\cite{jaynes1957information} as a guiding principle for synthetic tabular data, and introduce generative entropy maximization for tables---GEM-T---for this purpose. When only a limited set of expectations (e.g., marginal means, variances, and categorical frequencies) is known, the MaxEnt distribution is definitionally the least-biased model consistent with those constraints, maximizing the Shannon entropy of the resulting distribution. This approach explicitly avoids introducing additional bias into the synthetic data, and at the same time results in relatively compact models: GEM-T models have one visible layer and no hidden layers. MaxEnt (and therefore GEM-T) can be viewed as an extension, generalization, and refinement of Gaussian copulas that avoids some of its drawbacks, better capturing long tails and categorical variables. We demonstrate that GEM-T's performance matches ($n=2$) or exceeds ($n=21$) that of the state‑of‑the‑art synthetic tabular generators CTGAN \cite{ctgan} and TabularARGN (``TARGN'') \cite{tiwald2025tabularargn} on 23 of 34 widely used benchmark datasets (68\%).

\section{Related Work}\label{sec:related}

\subsection{Maximum-Entropy (MaxEnt) and Related Models}

MaxEnt has been applied across many fields, including neuroscience \cite{schneidman2006weak}, immunology~\cite{mora2010maximum, Arora519108}, reinforcement learning \cite{ziebart2008maximum}, and ecology \cite{PHILLIPS2006231}. MaxEnt models are a special case of the larger class of energy-based models \cite{Song:2021}, which define an energy function and thereby a probability distribution over all possible data values. When the energy function is constrained to be the minimal one that exactly reproduces a prescribed set of low‑order statistics (e.g., marginal means or pairwise correlations) while remaining otherwise maximally unbiased, the model becomes MaxEnt (see also Section \ref{sec:theory}). Other energy-based models include restricted Boltzmann machines, which introduce hidden binary units whose weights are learned through stochastic gradient methods \cite{smolensky1986information, hinton2002training}, and Hopfield networks, which employ symmetric deterministic connections to store attractor patterns \cite{hopfield1982neural}. Neither of these architectures enforces the explicit moment‑matching constraints that characterize MaxEnt models.

In the context of tabular data, MaxEnt has been used to a limited extent when the columns are purely categorical \cite{wu2016generatingrealisticsyntheticpopulation}, a major difference to the current work. Copulas \cite{nelsen2007introduction} and Bayesian networks  \cite{pearl1988probabilistic} are energy-based models that synthesize tabular data by fitting a parametric family of distributions and sampling from it. These are considered older or more classical approaches. As mentioned above, copulas are designed to capture dependencies between pairs of variables but often struggle with tail dependence by systematically underestimating joint extremes. Moreover, copulas generally require continuous marginal distributions; discretizing categorical variables therefore produces hard‑edge artifacts and biased dependence estimates. Bayesian networks use a graphical approach that can learn more complex conditional distributions. The Synthetic Data Vault (SDV) provides an implementation of the Gaussian copula \cite{patki2016sdv}, and PrivBayes \cite{zhang2017privbayes} generates private synthetic data from a Bayesian Network with noise injection.

\subsection{Deep Models}

Most contemporary generative models rely on DNN architectures. GANs \cite{goodfellow2014generative} have become popular because they achieved remarkable results on large‑scale image datasets. A GAN trains two networks simultaneously---a generator that creates synthetic samples and a discriminator that tries to separate real from synthetic data---using an adversarial loss that rewards the discriminator for correct classification and the generator for fooling it. Training is usually stopped when the discriminator’s accuracy on a held‑out validation set falls to chance level, indicating that it can no longer reliably tell real and generated examples apart. For tabular synthesis, the most widely used GAN is CTGAN \cite{ctgan}. An alternative to GANs is the variational auto‑encoder (VAE) \cite{kingma2019introduction}, which overcomes a key limitation of classic auto‑encoders for generative use: the latent space produced by a vanilla encoder is not regularized and therefore does not follow a known probability distribution. Instead of treating the latent space as a deterministic compressed representation, a VAE interprets each latent dimension as the parameters (mean and variance) of a simple prior distribution---commonly a standard normal---so that new samples can be generated simply by drawing a latent vector from that distribution and decoding it. A VAE variant designed for tabular data, TVAE, was introduced alongside CTGAN; however, CTGAN provides stronger privacy guarantees as its generator model does not have access to the training data. \cite{ctgan}.

Other notable deep‑neural‑network‑based generators include TARGN, which adopts an autoregressive modelling strategy \cite{tiwald2025tabularargn}. TARGN first discretizes all numerical columns, converting every column into a categorical variable, which by construction gives it strong handling of categorical data. Normalizing‑flow models \cite{dinh2016density} employ a neural network to learn an invertible transformation that maps a standard Gaussian latent variable to the data distribution of the training set. A specific variant, copula flow \cite{kamthe2021copula}, trains a neural network to learn a copula rather than the full joint probability density. All aforementioned methods train on a single dataset. In contrast, TabPFN---a classifier, not a generative model---is a foundation model that was pre‑trained on multiple tabular datasets to capture cross‑dataset patterns; it performs well on tables with fewer than 1,000 rows \cite{ye_closer_2025}. TabPFGen \cite{ma2024tabpfgen} converts TabPFN into a generative model by applying a joint energy‑based modelling framework. Finally, TabDDPM models tabular data with diffusion models \cite{kotelnikov2023tabddpm}.

\subsection{Quality Evaluation}\label{sec:quality}

Evaluating the quality of tabular synthetic data is inherently challenging. A widely adopted evaluation framework partitions the assessment into three dimensions: statistical resemblance, utility, and privacy~\cite{hernadez2023synthetic}. Statistical resemblance quantifies how closely the synthetic distribution matches the real one; it is generally regarded as the most objective of the three dimensions since it is independent of the task to which the synthetic data will be applied later. Common similarity metrics include the Kolmogorov–Smirnov test, earth‑mover’s (Wasserstein) distance, Jaccard score, and Kullback‑Leibler (KL) divergence; the recently introduced Eden score \cite{nguyenGradeInflationGenerative2024} addresses shortcomings of these traditional measures and better matches human evaluation of similarity between pairs of two-dimensional distributions.

Utility (or functional) tests typically involve training machine‑learning models on the synthetic data and then evaluating their performance on the original downstream task. These tests are useful when statistical resemblance alone does not reflect the requirements of a particular application; however, no single utility test can serve every purpose because the appropriate test depends on the downstream use case.

Evaluation for privacy is especially important when the source records contain sensitive information, a common situation in healthcare and demographic datasets, and is a major source of interest in generative models for tabular data. Privacy risk is often quantified using the distance to the closest record (DCR) \cite{Park_2018} and Jensen‑Shannon divergence (JSD) \cite{Joshi03042019}, both of which capture the likelihood of re‑identification, distributional leakage, or inadvertent memorization/reproduction of exact training data by the model.

\subsection{Our Contribution}

Our contribution in this work is twofold:
\begin{itemize}
  \item We introduce GEM-T, a MaxEnt approach that generalizes Gaussian copulas to higher dimensions and improves on some of their shortcomings, and
  \item We provide a comprehensive statistical empirical evaluation that demonstrates that GEM-T matches or outperforms recent DNNs on 23 of 34 common benchmark datasets.
\end{itemize}

\section{Methods}\label{sec:methods}

\subsection{Overview}

In tabular datasets, each row corresponds to an observation or datapoint and each column corresponds to a variable or feature. To synthesize data from an arbitrary underlying distribution, GEM-T learns a MaxEnt model that is constrained by a chosen set of feature functions of the variables, applying suitable pre-processing to improve the quality of the fit. A natural choice for the features are the distribution's mathematical moments: for example, the first‑order moment (mean), the second‑order moment (variance), and the third‑order moment (skewness) for every individual column, together with second‑ and third-order cross‑moments (covariances and co‑skewnesses) that capture relationships between columns. The user may decide the maximum order to fit; we typically fit second- and fourth‑order (kurtosis) moments (third-order tending to offer marginal improvement over second-order, while introducing mathematical inconveniences that fourth-order models lack). GEM-T then draws samples from the resulting MaxEnt fit, yielding synthetic data that resembles the original distribution.

\subsection{Datasets}

We tested on the 34 most-viewed datasets from the University of California-Irvine Machine Learning (UCIML) repository \cite{kelly2024uci} at the time of writing. This number is after excluding datasets with time series or natural-language entries, which were beyond the intended use case. We used the $lucie$ Python package \cite{ge2024textitlucieimprovedpythonpackage}. To ensure modelling efficiency and statistical power of results, we removed ID columns and aggregated  rare labels (those with $<$30 samples per label) into a single ``rare'' category in categorical columns. If a column had only a single value, or if it had only two values, one of which is ``rare'', and ``rare'' had $<$30 rows, the column was removed.

\subsection{Theoretical Background}\label{sec:theory}

\subsubsection{MaxEnt for Second-Order Moments}

Gaussian copulas capture joint dependence by first applying a marginal‑wise quantile (inverse‑CDF) transform to map each variable onto a standard normal, then fitting a multivariate Gaussian characterized by a mean vector and covariance matrix (from which standard deviations and pairwise covariances are derived), and finally applying the inverse transform to return to the original marginal distributions. Consequently, the Gaussian copula is a special case of a MaxEnt model that enforces only first‑ and second‑order moment constraints, because the normal distribution uniquely maximizes Shannon entropy given fixed means and covariances. Second‑order (i.e., covariance‑constrained) MaxEnt models admit closed‑form analytical solutions, whereas higher‑order MaxEnt models generally do not. This yields two practical advantages: (a) speed, because fitting a second‑order MaxEnt model requires no gradient descent or other optimization routine, and (b) sampling, because draws from a multivariate normal are computationally straightforward, while higher‑order distributions typically demand additional techniques such as Markov‑chain Monte Carlo. Second‑order (mean‑ and covariance‑constrained) fits are expected to perform well when the transformed data approximately follows a multivariate normal distribution. Conversely, absent additional considerations, departures from normality such as skewness, multimodality, or heavy tails are expected to degrade the quality of the fit.

\subsubsection{MaxEnt for Higher-Order Moments}

Fitting third- and fourth-order moments enable capture of some of these deviations; the additional constraints enable more precise models of the training data. It can be shown using Lagrange multipliers that MaxEnt probability distributions take the general form
\begin{equation}\label{maxent_features}
p_{\text{MaxEnt}}{(\textbf{x})} = \frac{1}{Z} \exp{\left( \sum_{i} \lambda_{i} f_{i}{(\textbf{x})} \right)}
\end{equation}
for some coefficients $\lambda_{i}$ (the weights of the model), features $f_{i}{(\textbf{x})}$ which we set to be the moments of the target distribution, and a normalizing constant $Z$. We can determine the weights $\lambda_{i}$ by requiring that the training data have the same expectation values of the features as the synthetic data:
\begin{equation}\label{moment_constraints}
\langle f_{i} \rangle_{model} \equiv \int p_{\text{MaxEnt}}{(\textbf{x})} f_{i}{(\textbf{x})} d\textbf{x} = \langle f_{i} \rangle_{real}
\end{equation}
In practice, this integral can rarely be evaluated analytically. Therefore the standard training method for MaxEnt models is to seek the values of the weights for which the training data has the maximum likelihood. A priori, this training method does not seem to incorporate information about the moment constraints (Eq. \ref{moment_constraints}). But a well-known and remarkable mathematical result shows that the gradient of the log-likelihood loss is the difference between expectation values of the features in the training data and in the synthetic data:
\begin{equation}\label{sample_gradient}
\frac{\partial}{\partial \lambda_{i}} \langle \log p_{\lambda}{(\textbf{x})} \rangle_{real} = \langle f_{i}{(\text{x})} \rangle_{real} - \langle f_{i}{(\textbf{x})} \rangle_{model}
\end{equation}
Thus at the optimum, the training data's expectation values are equal to those of the model, enforcing the moment constraints.

A drawback of the maximum-likelihood approach is that the expectation values must be estimated by Monte‑Carlo sampling at every training iteration. Consequently, training is bottlenecked by the sampler's speed, and the process introduces some noise into the gradient calculation. In practice, however, we still prefer this training method over sampling-free methods such as score-matching or noise-contrastive estimation \cite{song2021train, JMLR:v6:hyvarinen05a, pmlr-v9-gutmann10a}, because these alternatives are less directly aligned with the moment constraints than is the log-likelihood loss. As the order of the moments increases, the number of features grows combinatorially with the dimensionality of the target data (i.e., the number of columns), making training more challenging. We have found that using all moments up to fourth order---means, variances and covariances, skewnesses and coskewnesses, and kurtoses and cokurtoses---provides a favorable balance between expressive power and computational tractability. 

\subsection{Implementation}

\begin{figure}
    \centering
    \includegraphics[width=.4 \linewidth]{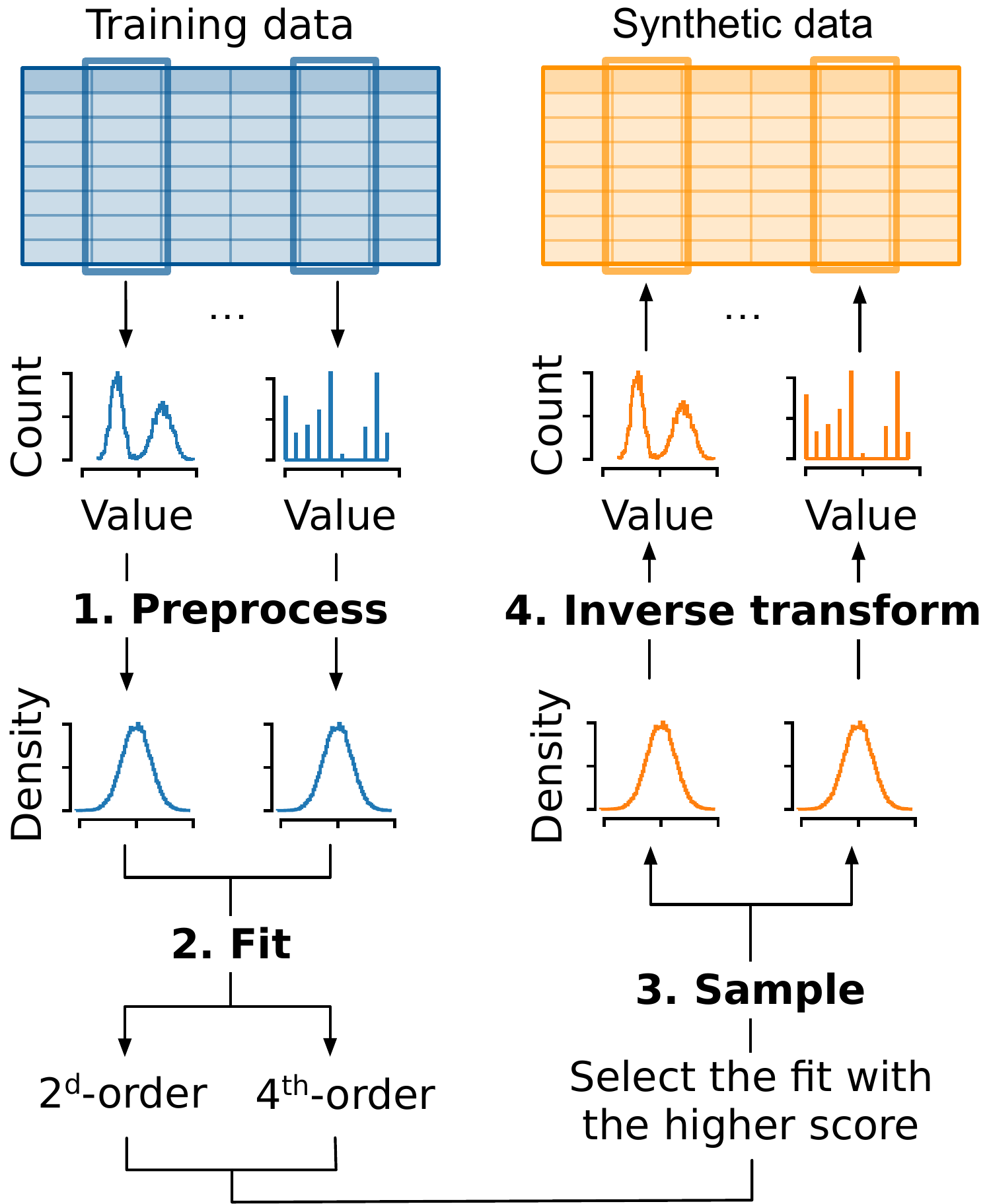}
    \caption{Overview of GEM-T. Numbering as in main text. Raw tabular training data from which histograms of individual columns are preprocessed/transformed by jittering,  quantile transformation, and scaling to normalize the data in transformed space. A second-order fit is computed analytically and a higher-order fit is performed iteratively via gradient descent, with moments as features, and the better model is chosen. Samples are drawn from the multivariate distribution, using Monte Carlo sampling for the higher-order fit (as during training epochs). Finally, samples are transformed back to the raw space by inverting the methods used in Step 1, resulting in generated synthetic data.}
    \label{fig:overview}
\end{figure}

Given a dataset with $n$ rows (datapoints) and $m$ columns (features), GEM-T generates synthetic data in four steps (Fig. \ref{fig:overview}):

\begin{enumerate}
\item Preprocess the dataset (the training data), transforming the marginals as appropriate
\item Fit a probability distribution to the transformed features
\item Draw synthetic samples from the fitted distribution (directly for second‑order fits or indirectly via Monte Carlo sampling for higher-order fits)
\item Apply the inverse marginal transformation to the sampled points (to render the generated synthetic data in the original/native data space).
\end{enumerate}

Details are described below.

\subsubsection{Data Preprocessing}

Preprocessing followed standard approaches of normalization and quantile transformation while providing novel adaptations to make these more applicable across binary, categorical, and continuous variable types. It proceeded in four steps (Fig. \ref{fig:overview}).

\textbf{1. Integer-encode categorical columns.} A column is considered categorical if it contains string inputs or if it contains no more than six unique values. Categories are integer-encoded. 

\textbf{2. Drop single-valued and nearly single-valued columns.} Columns consisting only of a single category---single-valued columns---are dropped as being too simple to require modeling. Any category occurring fewer than 30 times is considered too rare to model (insufficient statistical power) and is tagged as ``rare;'' if a column consists of only a single non-rare category and 30 or fewer rows tagged as rare, the column is considered nearly single-valued, and also dropped.

\begin{figure}
    \centering
    \includegraphics[width=1. \linewidth]{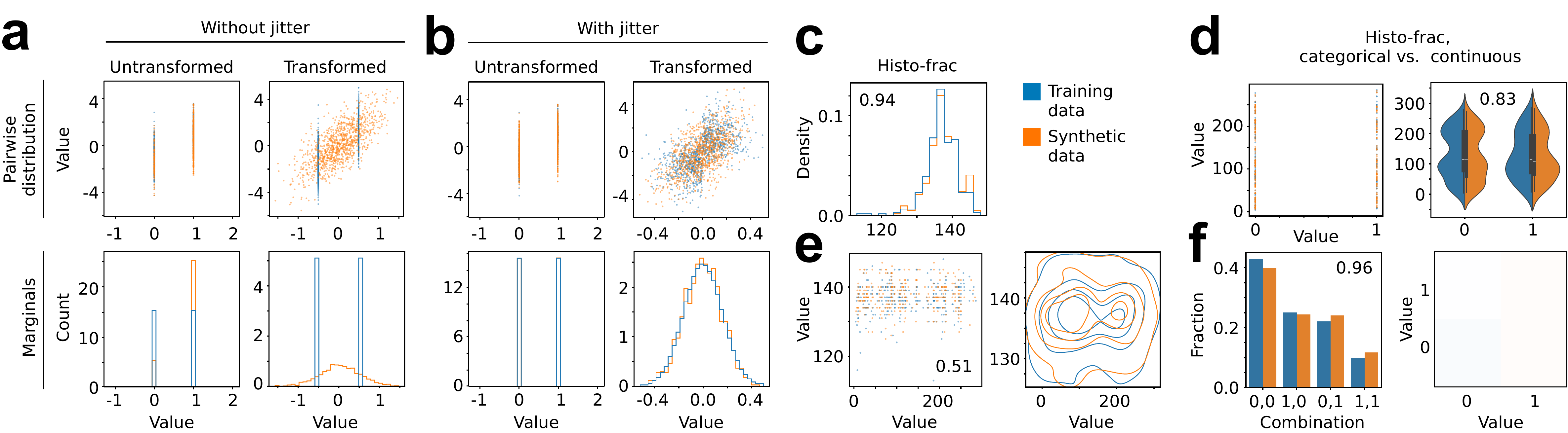}
    \caption{Preprocessing and evaluation: \textbf{a}) Left: Quantile transformation without jittering. Synthetic data fail to match the real data in both transformed and native space. Right: Quantile Transformation with jittering. By adding small random noise, quantile transformation is able to transform the distribution to an approximate normal. When transformed back to the native space, the binary distribution is successfully modeled. \textbf{b}) Evaluation measures. To examine how similar the distribution is for each column: histo-frac (row 1, left). To evaluate pair-wise relationships: Eden score for two continuous columns (row 2, left); average of each category's histo-frac for a pair with one continuous column and one categorical column (row 1, right); bar-frac for two categorical columns (row 2, right).}
    \label{fig:process-eval}
\end{figure}

\textbf{3. Normalize each column.} The approach is inspired by quantile transformation, the first step in the Gaussian copula method, but adds two important improvements. First, the data is jittered before the transform is applied. This is done by adding very small random noise to each value, of a magnitude far smaller than the minimum difference between values. The purpose is to impose a strict ordering by eliminating identical values, without which the quantile transform can fail (Fig. \ref{fig:process-eval}a right). In contrast, jitter enables it to perform a smooth mapping from the data to the normal (Fig. \ref{fig:process-eval}a left). A smooth normal makes sampling simpler. Second, we use the empirical cumulative distribution function (ECDF) to perform the probability integral transform. This is in contrast to existing Gaussian Copula synthesizers such as SDV's, which use either the CDF of some prior distribution (e.g., the beta distribution) or a KDE fit of the columns. While the KDE option of SDV is similar in spirit in that it is also non-parametric, the ECDF is less computationally demanding\footnote{In general, evaluating the KDE scales quadratically with the number of rows.}.

\textbf{4. Apply a modified min-max scaler.} This scales the quantile-transformed data to an interval of length $1$ such that the mean of the column is 0. This makes the first moments zero, offloading the work of fitting them to preprocessing. 

\subsubsection{Optimizer}

While second-order fits have a closed-form solution, higher-order fits require a different approach. We employ gradient descent. All moment weights are initialized to zero, meaning the first synthetic sample is drawn uniformly. The moments are measured, the gradient is calculated (Eq. \ref{sample_gradient}), the weights are updated, and the next sample is drawn; this process is repeated until a stopping condition is reached.

We use a hybrid implementation of the RProp \cite{riedmiller1993direct} and Adam \cite{kingma2014adam} optimizers to multiplicatively scale rates for each moment while simultaneously using momentum to stabilize the fit for noisy gradients. For RProp, we compare the signs of the first moment vector of Adam to its value from the previous epoch. For every moment with the same sign as an earlier epoch, we multiply its learning rate by a factor of $\gamma$. For every moment that switches signs, we decrease its learning rate by a factor of $\delta \times \gamma$; $\delta$ is set to be $>1$ as sign changes in an exponential moving average are more meaningful. We then update all moment weights with Adam, scaled by these individual rates.

We find that this approach converges faster than a pure Adam implementation, owing to its exponential rate growth, while effectively handling noisy gradients. We summarize the algorithm of our optimizer in the pseudo-code below. Table \ref{tab:optimizer_hyperparams} lists all hyperparameters used.

\begin{algorithm}[ht]
\caption{Adam--RProp Hybrid Optimizer}
\label{alg:adam-rprop}
\begin{algorithmic}[1]
\Require learning rate $\alpha>0$, decay rates $\beta_1,\beta_2\in[0,1)$, growth factor $\gamma>0$, deceleration factor $\delta>0$; initialize $\mathbf{w}_0$, $\mathbf{m}_0=\mathbf{0}$, $\mathbf{v}_0=\mathbf{0}$, $\mathbf{r}_0=\mathbf{1}$
\For{$t=1,2,\dots$}
  \State $\mathbf{g}_t \gets \; \langle f(\mathbf{x}) \rangle_{\text{real}} \;-\; \langle f(\mathbf{x}) \rangle_{\text{model}}$
  \State $\mathbf{a}_t,\mathbf{m}_t,\mathbf{v}_t \gets \textsc{Adam}\!\left(\mathbf{g}_t,\mathbf{m}_{t-1},\mathbf{v}_{t-1}\right)$
  \If{$\text{signChange}_t > \mathbf{0}$}
    \State $\text{rateUpdate}_t \gets 1+\gamma$
  \ElsIf{$\text{signChange}_t < \mathbf{0}$}
    \State $\text{rateUpdate}_t \gets (1+\gamma)^{-\delta}$
  \Else
    \State $\text{rateUpdate}_t \gets 1$ \Comment{no change if zero}
  \EndIf
  \State $\mathbf{r}_t \gets \mathbf{r}_{t-1} \odot \text{rateUpdate}_t$ \Comment{RProp-style step scaling}
  \State $\mathbf{w}_{t} \gets \mathbf{w}_{t-1} - \alpha \, (\mathbf{r}_t \odot \mathbf{a}_t)$ \Comment{parameter update}
\EndFor
\end{algorithmic}
\end{algorithm}

\renewcommand{\arraystretch}{1.2}
\begin{table}[ht!]
    \centering
    \begin{tabular}{|c|c|}
        \hline
        \textbf{Hyperparameter} & \textbf{Value} \\
        \hline
        learning rate & $1/\sqrt{n_\text{parameters}}$ \\
        growth factor & .005 \\
        deceleration factor & 6.5 \\
        Adam decay rate $\beta_1$ & .9 \\
        Adam decay rate $\beta_2$ & .999 \\
        \hline
    \end{tabular}
    \caption{Optimizer hyperparameter settings}
    \label{tab:optimizer_hyperparams}
\end{table}

\subsubsection{Sampler}

When fitting higher-order moments, to generate samples for per-epoch gradient calculation and final output, we sample from the non-normalized moment-based energy distribution defined by the model to that point (at a given epoch or the final trained model), while avoiding low-energy artifacts outside the support of the training data that occur due to the diverging nature of odd-ordered moment terms.

Sampling is performed using a Metropolis-Hastings sampler bounded by an ellipsoid calculated from the original data's covariance matrix (with Mahalanobis distance squared cutoff = 25). Metropolis-Hastings sampling uses Markov chain Monte Carlo to collect a sequence of data points sampled according to the energy distribution. For each sampling chain, a random walk starts from the mean of the data distribution (when preprocessing with the min-max scaler, this is the origin), and a new point is randomly proposed for each step using the scaled covariance matrix of the original data. If this proposal falls outside the ellipsoidal bound, it is reflected back along the norm of the closest ellipsoid boundary point until it is valid. The proposal point is either accepted or rejected based on the calculated ratio of its energy and the energy of the previous point according to the Metropolis-Hastings acceptance criterion. This process is repeated until a sufficiently large sample is generated. A default 200 steps of burn-in were used to avoid out-of-distribution samples and thinning (keeping only every $k^\text{th}$ element; default $k=25$) was employed to minimize autocorrelation. 

\subsubsection{Applying constraints}
After the inverse transform is applied to return the data to its native space, constraints are applied to reflect certain features of the training data (``coercion''). In integer-type columns, including integer-encoded categories, sample values are rounded to the closest integer. Samples violating non-positivity/negativity constraints (inferred automatically from the training data) are dropped and replaced via additional sampling.

\subsubsection{Missing values}
In principle, GEM-T requires that all the moments up to the prescribed order can be calculated so no feature is missing. In practice, missingness is handled as follows. For categorical columns, missing values are given their own category. When pairs of continuous columns are disjoint, meaning they have no rows in which both columns are non-missing, covariance and higher-order moments cannot be calculated. In such cases, missing values in the covariance matrix are filled in following the maximum-entropy principle: we identify the normal distribution with the largest possible entropy using the non-missing entries in the covariance matrix as constraints; since the entropy of a multivariate Gaussian is a function of the determinant of the covariance matrix: 
\begin{equation}
    H = \frac{1}{2} \ln \left[ (2 \pi e)^k \, \mathrm{det}|\Sigma| \right]
\end{equation}
we fill with values that maximize the determinant. 

When the missingness pattern in the training data does not cause disjointness, it may still cause the covariance matrix to not be positive semi-definite. This may indicate that the missingness pattern is not sufficiently random that one can safely assume that the non-missing data is representative of all the data (missing and non-missing). In such cases we diagonalize the covariance matrix, take the absolute value of the eigenvalues, and undo the diagonalization. This approach mitigates potential numerical instability, which can cause a small positive eigenvalue to be represented as slightly negative, while giving reasonable results when eigenvalues are truly negative.

\subsection{Evaluation} 
Considering code availability and the ability of available code to deal with various data types, we benchmarked GEM-T against two other high-profile/state-of-the-art models: TARGN, an autoregressive network-based synthetic data generator developed by MOSTLYAI \cite{tiwald2025tabularargn}, and CTGAN, a GAN-based deep-learning synthesizer \cite{ctgan}.

\subsubsection{Complexity}
The number of parameters for GEM-T's second- and fourth-order models scales with the number of columns $n$ as $n^2+n/2$ and $(1/24)n^4 + (5/12)n^3 + (35/24)n^2 + (25/12)n$, respectively. For a 14-column table, the median across our datasets (Table \ref{tab:appendix}), these equate to 203 and 3,059 parameters, respectively.

\subsubsection{Runtimes and stopping conditions}
GEM-T first calculates the second-order solution, which is quick because it can be done analytically. It then attempts to boost performance by performing a higher- (here, fourth-)order fit with two stopping conditions: a training time of six hours (executed on a 96-Core AMD Ryzen Threadripper PRO 7995WX with two threads per core; 70 threads utilized for each run) or when the average quality score in the last $n$ epochs (here, 400) stops improving, whichever comes first. To account for sampling variance, we pick the weights from the epoch of best training performance in the last $2n$ epochs and sample with the weights to produce the final output of synthetic data. The model at this epoch is the final model used for sampling.

TARGN has an early stopping mechanism that halts training when validation loss stops decreasing. CTGAN does not incorporate early stopping, nor does it support restarting fits, but does have a parameter for setting the maximum number of epochs $n_\text{max}$. In our testing, it usually took CTGAN very many epochs to achieve decent performance; for a fair comparison to GEM-T, we estimated CTGAN's time-per-epoch as the average across 10 epochs, divided this time into six hours (the time limit for GEM-T), and set $n_\text{max}$ accordingly.

\subsubsection{Quality}
To measure the quality of synthetic data, we focus on two aspects: statistical similarity and privacy protection. 

Statistical similarity is scored by how closely distributions drawn from the synthetic data match the training data. Because this is difficult in higher dimensions, we measure how well the distributions match for each column and each pair of columns, yielding 1- and 2-dimensional quality scores, respectively, which each range from 0 to 1 (with 1 meaning the a perfect match to the training data). We then average together all the 1-dimensional scores, separately average the 2-dimensional scores, and finally average these averages to yield our final statistical similarity score. This stepwise averaging is to place the 1- and 2-dimensional scores on an even footing, since the $n(n-1)/2$ 2-dimensional scores greatly outnumber the $n$ 1-dimensional scores.

\subsubsection{1D quality scores for single-column distributions} 
For each column in the dataset, we create histograms for both the real and synthetic datasets. Because many similarity measures suffer from ``grade inflation'' \cite{grade_inflation}, yielding unreasonable values, we compared histograms using a ``histo-frac'' score, which captures the difference between the density histograms. To do this, we break the total range according to convention into $\sqrt{n_\text{rows}}$ bins (when there are at least 10 unique values) or no.-unique-values bins (otherwise), sum the (absolute value of the) per-bin differences, normalize, and then subtract from 1 to get a similarity:

\begin{equation}
\text{histo-frac score} = 1 - \frac{1}{2} \sum_{i} |h_{i}^\text{real} - h_{i}^\text{synthetic}|
\end{equation}

where $h_{i}^\text{real}$ and $h_{i}^\text{synthetic}$ are the $i^\text{th}$ bin counts of the real histogram and of the synthetic histogram, respectively. 

\subsubsection{2D quality scores for pairwise distributions}
Evaluation depends on whether the two columns are continuous, discrete, or a combination (Fig \ref{fig:process-eval}b): 

\begin{itemize}
\item When both columns are continuous, we use the Eden score (which avoids grade inflation) \cite{grade_inflation}. In (very) rare cases, certain columns are highly imbalanced, consisting predominantly of a single value with only a few rare alternatives. After applying KDE's threshold, these infrequent values may be too few to support meaningful Eden score calculation and scores for these pairs are not recorded (affected datasets include ``Spambase'' and ``Forest Fires'').

\item When both columns are discrete, we calculate the ``bar-frac'' score, which is essentially a matrix subtraction. For example, when both columns are binary,  we compute the frequencies of all 4 possible values---(0,0), (0,1), (1,0), and (1,1)---for both the real data and the synthetic data, perform a subtraction to get the difference, normalize, and subtract from 1. This is the same as treating these four as bins and calculating the histo-frac score. The differences are visualized using heatmaps in which the deeper the color, the larger the gap.

\item When one column is discrete and the other continuous, the score is the average of histo-frac scores for all unique values in the discrete columns. This can be visualized using violin plots, comparing the distributions for real and synthetic data within each class.
\end{itemize}

For speed, we perform these calculations on a maximum of 5,000 datapoints and plot accordingly.

\subsubsection{Indistinguishability and overfitting}\label{overfitting}

Scoring too highly suggests overfitting, as synthetic data is not expected to match the training data perfectly. However, what ``too highly'' means might vary from dataset to dataset. As a rule of thumb, synthetic data should not fit the training data better than training data fits itself; the latter provides a dataset-specific benchmark. We use this principle to test for overfitting as follows. We randomly split the training data in half, and score one half against the other as above. We do this multiple times (50) to get a distribution of scores; the range reflects a best-case scenario for synthetic data (black bars in Fig. 3b). If the synthetic data's score falls within this range, it is considered statistically indistinguishable from the training data. This enables a p-value for the null hypothesis that the synthetic data is indistinguishable from the training data. Note that to be a fair comparison, in this test the number of rows sampled to create the synthetic data is the same as half the number of rows in the full training data.

\subsubsection{Privacy preservation}
DNNs can be prone to overfitting to input data, at the risk of leaking private records \cite{hayes2017logan}. The degree of privacy in the synthetic data is tested with the distance-to-closest-record (DCR) measure, using SDV's DCRBaselineProtection function. This is calculated using a combination of the absolute value and Hamming distances for continuous and categorical columns, respectively. Specifically, for each dataset, DCRBaselineProtection compares the median DCR of the synthetic data to the median DCR of a random set of data and outputs their ratio as a score. A score of 0 means perfect similarity to the training data, while 1 means a distance comparable to random noise. The same random data baseline was used for all models tested.

\section{Results} \label{result}

\begin{table}[tbp]
\centering
\begin{tabular}{p{4cm}|r|r||c|c|c||c|c}
\toprule
Dataset & Rows & Cols & CTGAN & TARGN & GEM-T & $2^\text{nd}$ & $4^\text{th}$ \\
\midrule
\hline
Abalone & 4177 & 9 & 0.73 & \textbf{0.80} & 0.79 & 0.76 & \textit{0.79} \\
Adult & 48842 & 15 & 0.85 & \textbf{0.91} & 0.88 & \textit{0.88} & 0.88 \\
Auto MPG & 398 & 8 & 0.73 & 0.52 & \textbf{0.77} & \textit{0.77} & 0.74 \\
Automobile & 205 & 24 & 0.73 & 0.62 & \textbf{0.79} & \textit{0.79} & 0.78 \\
Bank Marketing & 45211 & 17 & 0.89 & \textbf{0.94} & 0.91 & \textit{0.91} & 0.91 \\
Breast Cancer & 286 & 10 & 0.93 & 0.81 & \textbf{0.94} & 0.93 & \textit{0.94} \\
Br. Cancer Wisc. (Diag.) & 569 & 31 & 0.68 & 0.70 & \textbf{0.83} & 0.81 & \textit{0.83} \\
Br. Cancer Wisc. (Orig.) & 699 & 10 & 0.60 & 0.63 & \textbf{0.80} & 0.78 & \textit{0.80} \\
Car Evaluation & 1728 & 7 & 0.93 & \textbf{0.96} & 0.96 & \textit{0.96} & 0.96 \\
CDC Diabetes Health Ind. & 253680 & 22 & 0.92 & \textbf{0.97} & 0.93 & \textit{0.93} & 0.92 \\
Chronic Kidney Disease & 400 & 24 & \textbf{0.77} & 0.69 & 0.77 & \textit{0.77} & 0.76 \\
Concrete Compressive Str. & 1030 & 9 & 0.69 & 0.71 & \textbf{0.73} & 0.73 & \textit{0.73} \\
Credit Approval & 690 & 16 & \textbf{0.89} & 0.85 & 0.88 & 0.88 & \textit{0.88} \\
Default Credit Card Clients & 30000 & 24 & 0.76 & \textbf{0.79} & 0.75 & \textit{0.75} & 0.72 \\
Diabetes 130-US Hospitals & 101766 & 39 & 0.87 & \textbf{0.91} & 0.88 & \textit{0.88} & 0.79 \\
Dry Bean & 13611 & 17 & 0.56 & 0.78 & \textbf{0.78} & 0.75 & \textit{0.78} \\
Energy Efficiency & 768 & 10 & 0.70 & \textbf{0.83} & 0.82 & 0.80 & \textit{0.82} \\
Estim. of Obesity Levels & 2111 & 17 & 0.82 & 0.87 & \textbf{0.89} & 0.88 & \textit{0.89} \\
Forest Fires & 517 & 13 & 0.75 & 0.64 & \textbf{0.82} & 0.82 & \textit{0.82} \\
Heart Disease & 303 & 14 & 0.86 & 0.74 & \textbf{0.88} & \textit{0.88} & 0.87 \\
Heart Failure Clinical Rec. & 299 & 13 & 0.84 & 0.73 & \textbf{0.90} & \textit{0.90} & 0.89 \\
Iris & 150 & 5 & 0.67 & 0.45 & \textbf{0.74} & 0.71 & \textit{0.74} \\
MAGIC Gamma Telescope & 19020 & 11 & 0.76 & \textbf{0.85} & 0.82 & 0.80 & \textit{0.82} \\
Mushroom & 8124 & 22 & 0.90 & \textbf{0.94} & 0.92 & \textit{0.92} & 0.92 \\
Online Shoppers Purchasing Intention & 12330 & 18 & 0.82 & \textbf{0.89} & 0.82 & 0.80 & \textit{0.82} \\
Predict Students' Academic Success & 4424 & 37 & 0.79 & 0.79 & \textbf{0.79} & \textit{0.79} & 0.74 \\
Real Estate Valuation & 414 & 7 & 0.66 & 0.63 & \textbf{0.75} & 0.69 & \textit{0.75} \\
Rice & 3810 & 8 & 0.68 & 0.76 & \textbf{0.78} & \textit{0.78} & 0.78 \\
Spambase & 4601 & 58 & 0.65 & 0.70 & \textbf{0.78} & \textit{0.78} & 0.50 \\
Statlog (Ger. Credit Data) & 1000 & 21 & 0.90 & 0.91 & \textbf{0.95} & \textit{0.95} & 0.95 \\
Student Performance & 649 & 33 & 0.89 & 0.89 & \textbf{0.94} & \textit{0.94} & 0.93 \\
Wholesale customers & 440 & 8 & 0.82 & 0.76 & \textbf{0.88} & \textit{0.88} & 0.83 \\
Wine & 178 & 14 & 0.70 & 0.46 & \textbf{0.71} & \textit{0.71} & 0.70 \\
Wine Quality & 6497 & 13 & 0.71 & 0.80 & \textbf{0.84} & 0.82 & \textit{0.84} \\
\bottomrule
\end{tabular}
\vspace{1em}
\caption{Model performances on the UCIML benchmark. Best results in \textbf{bold} and \textit{italics}.}
\label{tab:appendix}
\end{table}

\begin{figure}
    \centering

    \includegraphics[width=1\linewidth]{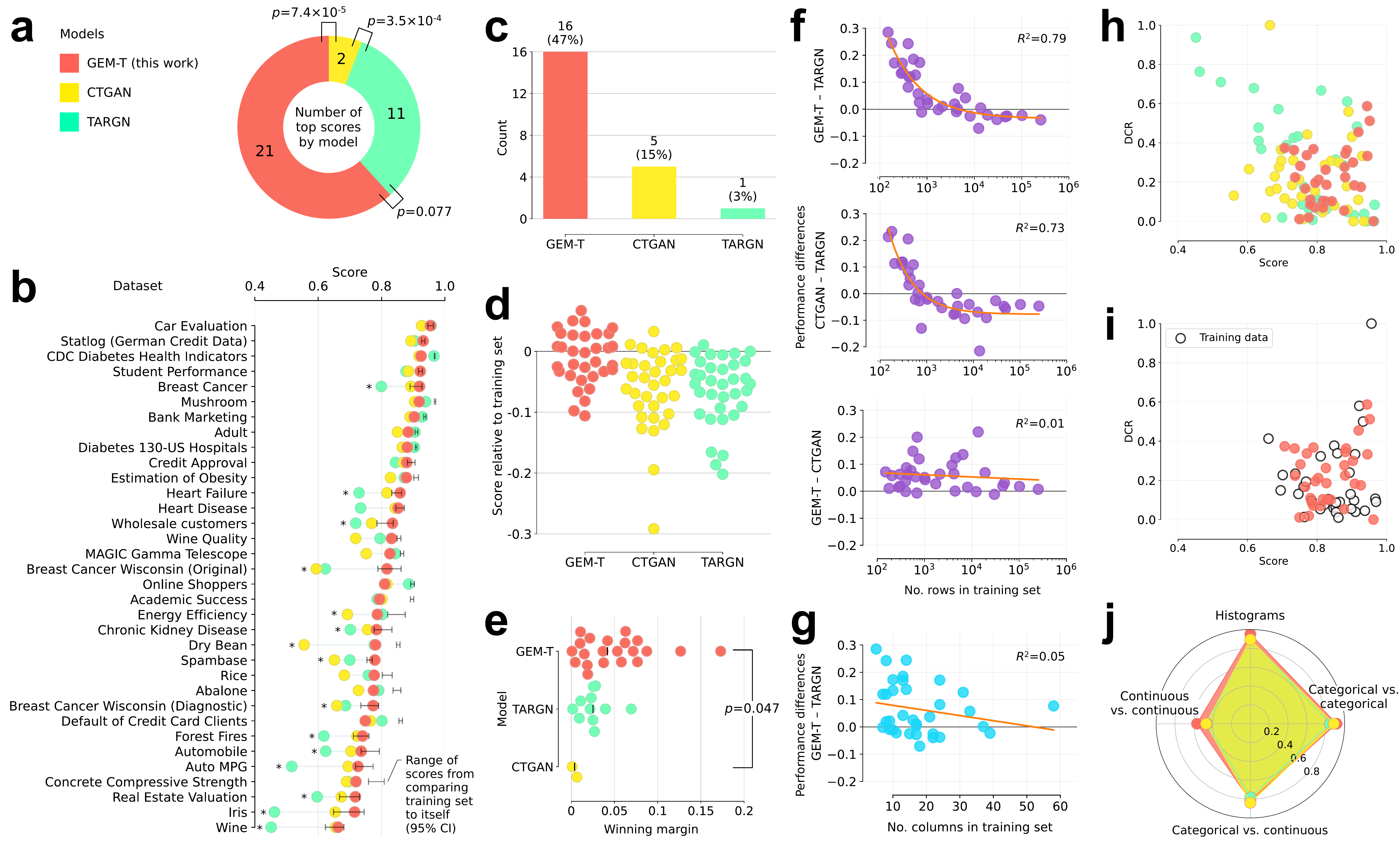}
    \caption{Performance of GEM-T vs. other methods. \textbf{a} Number of datasets on which each model achieves the top score. \textbf{b} Scores of each model on each dataset. Black bars show the 5$^\text{th}$-to-95$^\text{th}$ percentile range of scores obtained by taking a random half of each training set and comparing it to the other half; models in this range are indistinguishable from the training set. Asterisks indicate models that scored $\geq$0.1 worse than the best performer. \textbf{c}  The number of datasets on which each model reaches this range and \textbf{d} how close each fit comes to this range. \textbf{e} The difference in score between the best and second-best model (black ticks = medians) and how differences between models vary with \textbf{f} the number of rows in the training data and \textbf{g}  the number of rows in the training data. \textbf{h} Distance-to-closest-record (DCR) vs. score for each model on each dataset and \textbf{i} how synthetic data from GEM-T compares to a random half of the training data. \textbf{j} How well each model fits marginals (``Histograms'') and two-dimensional relationships between data columns, broken down by data type.}
    \label{fig:results}
\end{figure}

We trained GEM-T, TARGN, and CTGAN on the top 34 most-viewed datasets from the UCIML database and scored each fit for quality and privacy (Table \ref{tab:appendix}). The datasets covered a wide range of domains, including finance (Bank Marketing, Credit Approval) and health (Breast Cancer, Heart Disease), among others (Wine Quality, Auto MPG, Forest Fires).

GEM-T scored highest in 21 of the 34 datasets (61.7\%), including 4 of 7 financial datasets (57.1\%) and 7 of 9 healthcare datasets (77.8\%; Fig. \ref{fig:results}a). It consistently scored within or very near the reference range indicating indistinguishability from the training data, the exceptions being Default of Credit Card Clients, Academic Success, and Dry Bean, on which all of the models had difficulty (Fig. \ref{fig:results}b). The upper-95\% limit ranged considerably across datasets, from 0.681 (for Mushroom) to 0.971 (for Wine) (Fig. \ref{fig:results}b). Although the best scores for all models were generally close to these limits, ranging from 0.664 (for Wine) to 0.965 (for CDC Diabetes Health Indicators); CTGAN (5/34) and especially TARGN (10/34) sometimes scored substantially worse ($\geq$0.10) than the best model for a given dataset, which GEM-T never did (0/34) (Fig. \ref{fig:results}b, asterisks). Aside from the above three exceptions, synthetic data from the best-performing model was never further than 0.03 from the reference interval. In all, GEM-T's score reached the reference interval in 16 of the 34 datasets (47.1\%), compared with 5 for CTGAN and 1 for TARGN (Fig. \ref{fig:results}c-d).

GEM-T also had the highest winning margins, i.e. the differences between it and the second-highest-scoring model for the 21 datasets on which it scored highest (median, 0.042; Fig. \ref{fig:results}e). This includes not only the two Breast Cancer Wisconsin datasets, in which both TARGN and CTGAN particularly struggled, but other datasets such as Spambase and Heart Failure. TARGN's median winning margin was lower, at 0.025; its strongest relative performance was a margin of 0.069 for Online Shoppers. Meanwhile, CTGAN barely outperformed the others on the two datasets on which it scored highest, with a mean margin of only 0.004. GEM-T and TARGN performed similarly for large datasets but TARGN struggled with smaller ones, with a pronounced trend toward worse performance on datasets with under 10,000 rows (Fig. \ref{fig:results}f). This trend was well explained by an exponential, with an $R^2$ of 0.79 ($p=2.0\times10^{-12}$). Meanwhile, GEM-T outperformed CTGAN by a fairly consistent 0.05 independent of dataset size ($R^2$=0.01). There were no meaningful trends with the number of columns ($R^2<$0.05 and $p>$0.05 for all model comparisons, e.g. Fig. \ref{fig:results}g).
 
GEM-T's better performance came without sacrificing privacy relative to the other models, as assessed by DCR (Fig. \ref{fig:results}h). In fact, DCR for the synthetic data generated by GEM-T was no lower than for a random half of the training data compared against the other half (Fig. \ref{fig:results}i).

\begin{figure}
    \centering
    \includegraphics[width=1.\linewidth]
    {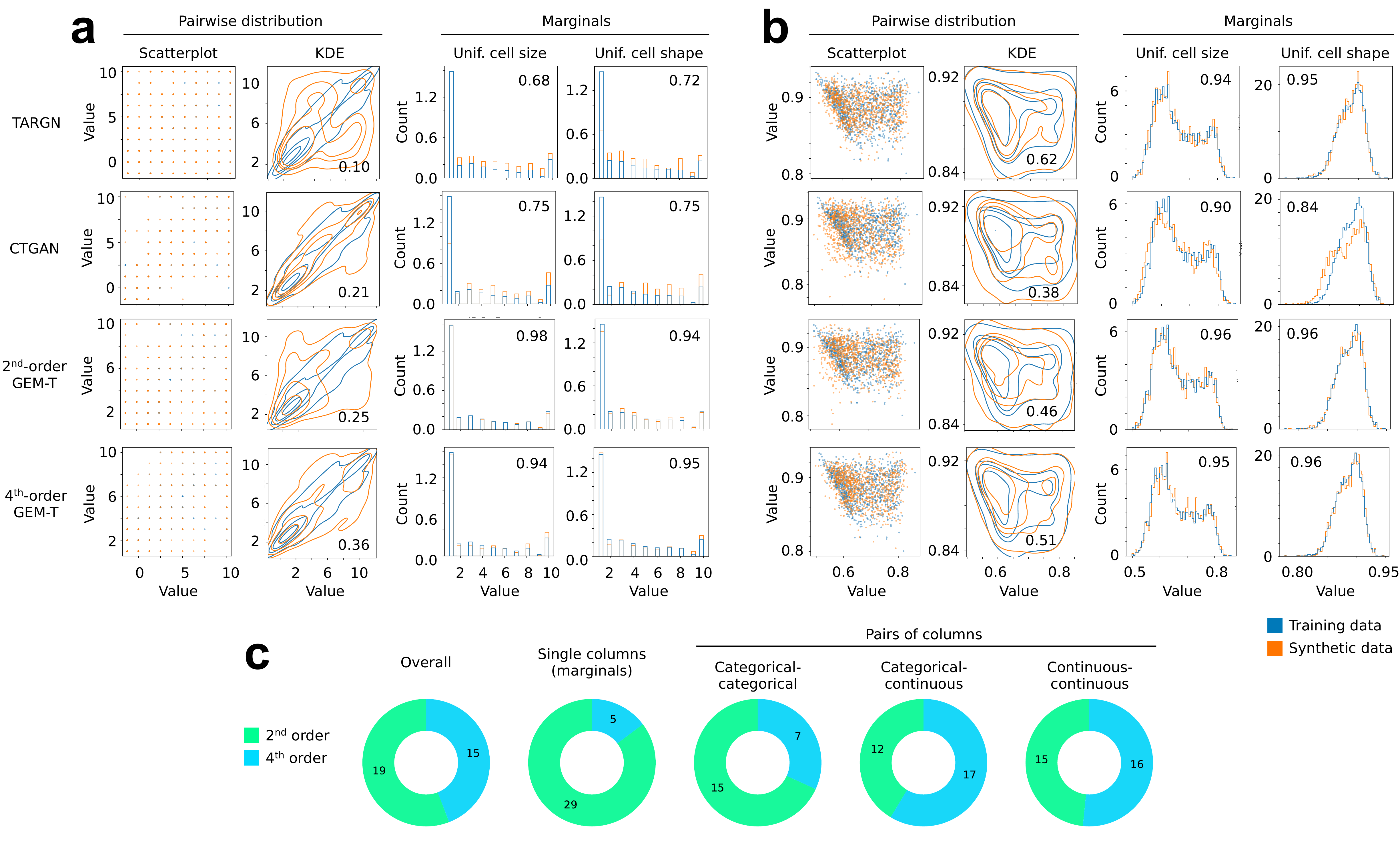}
    \caption{Comparison of marginal and pairwise fits for \textbf{a} two highly correlated integer columns from the Breast Cancer Wisconsin (Diagnostic) dataset and \textbf{a} two continuous columns with a hard-edge relationship from the Rice dataset (the diagonal edge at lower left). Numbers indicate histo-frac scores for the marginal distributions and Eden scores for the pairwise distributions.}
    \label{fig:dis}
\end{figure}

Comparing pairwise distributions of variables between the training data and synthetic data showed that GEM-T's higher scores were in part due to better handling of tails, hard edges, and other challenging features by higher-order moments (Fig. \ref{fig:dis}). For example, in the Breast Cancer Wisconsin (Diagnostic) dataset, the columns entitled ``Uniformity of cell size'' and ``Uniformity of cell shape''---both integer columns that take on values from 1 to 10---are highly correlated, such that most of the pairwise distribution lies along the diagonal. The synthetic data from CTGAN and especially TARGN have substantial off-diagonal density, resulting in low Eden scores of 0.21 and 0.10, respectively; GEM-T's second-order fit does little better, with an Eden score of 0.25 (Fig. \ref{fig:dis}a). In addition, TARGN and CTGAN have difficulty fitting the marginals (the 1-D distributions or histograms), with histo-frac scores of 0.68-0.75 vs. 0.94-0.98 for GEM-T's second-order fit. Note the excellent fit of the marginals by the second-order fit is insufficient to handle the tail.

In contrast, GEM-T's fourth-order fit has much less off-diagonal density, reflected by a substantially higher Eden score of 0.36 (as well as histo-frac scores of 0.94 and 0.95). This is also the case in continuous data, for example in the ``Area'' and ``MinorAxisLength'' columns of the Dry Beans dataset, in which CTGAN and the second-order fit struggle, TARGN performs better, and GEM-T performs best. The ``Eccentricity'' and ``Extent'' columns of the Rice dataset demonstrate the value of higher moments for fitting hard edges, which CTGAN and the second-order fit struggle with (Fig. \ref{fig:dis}b). Overall, second-order fits scored better on marginals ($\chi^2$ $p=3.9\times10^{-5}$) (Fig. \ref{fig:dis}c). However, marginals are greatly outnumbered by pairwise relationships, which fourth-order fits scored better on, especially those involving continuous columns. In all, the GEM-T fit was nearly equally likely to be second- vs. fourth-order (\textit{p} for difference = 0.49).

\section{Discussion}

GEM-T is a lightweight generative model based on the principle of MaxEnt, which seeks the least-biased model possible for a given set of data. It includes improvements on standard Gaussian copulas model that make it more suitable for tabular data, as well as a generalization to higher-order moments involving sampling and optimization for more complex datasets. We have demonstrated that GEM-T achieves very good statistical performance on a suite of public datasets, where it outperforms deep models in the majority of cases despite its much simpler architecture. Notably, GEM-T never scored uncharacteristically poorly, unlike the deep models we tested (Fig. \ref{fig:results}b, asterisks).

The benchmark suite included the top datasets in UCIML, a standard resource in the field, and spanned a broad range of domains, with nearly all datasets containing both numerical and categorical features. GEM-T performed well on both small, simple datasets (Iris) and large, high-dimensional ones (CDC Diabetes Health Indicators), showing it to be both robust and adaptable. GEM-T's solid performance even on smaller datasets is notable, since adding to or otherwise expanding on small datasets is a prime motivation for using generative models. We do note that UCIML's datasets are relatively clean compared to real-world data where additional challenges often arise, and also that we deliberately excluded time series or free-form text features, which are left for future work.

Our results clearly show that some datasets are harder to model than the others, as evidenced by a failure of any model to come close to the indistinguishability threshold in a few cases, specifically Academic Success and Dry Bean (Fig. \ref{fig:results}b). While is not obvious what makes these datasets more difficult, we suspect the data-specific characteristics has contributed to this. For example, several columns in Academic Success are integer-encoded with large gaps and arbitrary ordering (columns to do with occupation and nationality), even for categories that do appear to have a native order (columns to do with qualification, which includes number of years of schooling). However, we note that in every case where the threshold was reached, GEM-T was one of the models that did so (sometimes the only one), and that synthetic data from GEM-T tended to be closer to the threshold than the other models. In other words, GEM-T generated higher-quality synthetic data more often and more consistently than its deep-learning-based predecessors.

How can relatively small numbers of moment constraints outperform deep models? We believe an important contributor is the initial preprocessing, especially the quantile transform. This non-parametric step can be thought of as warping the native coordinate system to one in which the data is more regular for lower moments. There is reason to expect lower moments to contain more information and higher moments to provide refinements \cite{bialek2007rediscovering}, explaining the utility of such a transformation. Specifically, quantile transformation is helpful for fitting discrete, irregularly-spaced numerical columns, i.e., columns that take only a handful of values, with gaps between. For such columns, the quantile transform acts as a soft coercion: it ensures that the synthetic data in that column is highly concentrated near the values found in the training data. This soft coercion is only approximate, but it also facilitates the second layer of hard coercion in our implementation. Quantile transformation also facilitates fits of multimodality. As a thought experiment, consider fitting 1-dimensional training data with many modes, using a fourth-order fit. Without the quantile transform, the fit's energy function is a quartic polynomial, which can have at most two maxima, meaning it would only be able to fit two modes at most. By converting to a normal, the quantile transform makes it much easier to capture multimodality in the marginals. Meanwhile, higher-order moments mitigate the tail-dependence issue that quantile transformation can otherwise introduce (which Gaussian copulas suffer) (Fig. \ref{fig:dis}).

GEM-T combines second- and fourth-order fits to balance the value of stronger constraints against the speed and simplicity of analytical solutions. Internal to the model, second- and fourth-order fits each win about half the time, suggesting diminishing returns for higher moments, issues with sampling variance and the robustness of gradient descent, or both. Diminishing returns are consistent with prior observations that pairwise correlations already contain sufficient information to generate sequences of new functional proteins \cite{bialek2007rediscovering}. Additionally, the gradient descent algorithm becomes more sensitive and time-consuming as the number of columns in a table increases, since the number of features scales super-linearly with the number of columns, so a performance increase becomes harder to achieve for larger datasets. Internally to GEM-T, the near-tie of second- vs. fourth-order fits, and the difference specifically in marginals (Fig. \ref{fig:dis}c), suggests that principled selection of subsets of features might lead to still-better results. This could be preferable when memory and/or training time is limited. Investigating criteria for selecting such subsets is an interesting avenue for future research.

We note three limitations. First, we measured univariate and pairwise statistical fidelity but not higher-order relations, even as we consider the evaluation metrics we used to be more stringent than other common methods like correlation similarity and Earth Mover's distance (see Section \ref{sec:quality}). Second, we note that high statistical fidelity may not necessarily guarantee higher performance in downstream tasks like classification and regression \cite{hansen2023reimaginingsynthetictabulardata}, However, these are difficult to assess as the number of such tasks is effectively unbounded. They are therefore left for future work. A third limitation relates to the scaling behavior of this first implementation of GEM-T, as the number of features increases rapidly with the number of columns and with the order of the fit. As mentioned, comparison of results from second- vs. fourth-order fits suggests there is useful optimization in this direction.

\section{Conclusion}

We have demonstrated that GEM-T is a powerful generator of synthetic tabular data, especially when data is limited. On a comprehensive benchmark spanning many datasets, it achieves higher statistical quality scores in a strong majority of cases while maintaining a reasonable measure of privacy. This result underscores the robustness and flexibility of remarkably simple moment-driven energy-based models, while highlighting the fact that aggregating moments over expectations is an effective way to prevent the memorization of training data. 

\section{Acknowledgements}
This work was supported by the National Institutes of Health (R01HL150394, R01HL150394-SI, R01AI148747, and R01AI148747-SI), the Gordon and Betty Moore Foundation, the Food and Drug Administration, and the Tianqiao and Chrissy Chen Institute.

\bibliographystyle{unsrt}
\bibliography{bibliography}

\end{document}